\documentclass[runningheads]{llncs}

\usepackage{amssymb}
\usepackage{graphicx}

\usepackage{url}
\urldef{\mailsa}\path|{alfred.hofmann,ursula.barth,ingrid.beyer,natalie.brecht,|
\urldef{\mailsb}\path|christine.guenther,frank.holzwarth,piamaria.karbach,|
\urldef{\mailsc}\path|anna.kramer,erika.siebert-cole,lncs}@springer.com|

\begin{document}

\mainmatter  

\title{Probabilistic Inferences in Bayesian Networks}

\titlerunning{Probabilistic Inferences in Bayesian Networks}

\author{Jianguo Ding}

\authorrunning{Jianguo Ding}

\institute{Interdisciplinary Center for Security, Reliability and Trust\\
University of Luxembourg,
Luxembourg\\
\email{Jianguo.Ding@ieee.org}\\
} \maketitle

\begin{abstract}
Bayesian network is a complete model for the variables and their
relationships, it can be used to answer probabilistic queries about them. A Bayesian network can thus be considered a
mechanism for automatically applying Bayes' theorem to complex problems.
In the application of Bayesian networks, most of the work is related to probabilistic
inferences. Any variable updating in any node of Bayesian networks
might result in the evidence propagation across the Bayesian networks.
This paper sums up various inference techniques in Bayesian networks
and provide guidance for the algorithm calculation in probabilistic inference in
Bayesian networks. 
%
\end{abstract}

\section{Introduction}
Because a Bayesian network is a complete model for the variables and
their relationships, it can be used to answer probabilistic queries
about them. For example, the network can be used to find out updated
knowledge of the state of a subset of variables when other variables
(the evidence variables) are observed. This process of computing the
posterior distribution of variables given evidence is called
probabilistic inference. A Bayesian
network can thus be considered a mechanism for automatically
applying Bayes' theorem to complex problems.

In the application of Bayesian networks, most of the work is related
to probabilistic inferences. Any variable updating in any node of
Bayesian networks might result in the evidence propagation across
the Bayesian networks. How to examine and execute various inferences
is the important task in the application of Bayesian networks.

This chapter will sum up various inference techniques in Bayesian
networks and provide guidance for the algorithm calculation in
probabilistic inference in Bayesian networks. Information systems are of discrete
event characteristics, this chapter mainly concerns the inferences in discrete events
of Bayesian networks.

\section{The Semantics of Bayesian Networks}

The key feature of Bayesian networks is the fact that they provide
a method for decomposing a probability distribution into a set of
local distributions. The independence semantics associated with
the network topology specifies how to combine these local
distributions to obtain the complete joint probability
distribution over all the random variables represented by the
nodes in the network. This has three important consequences.

Firstly, naively specifying a joint probability distribution with
a table requires a number of values exponential in the number of
variables. For systems in which interactions among the random
variables are sparse, Bayesian networks drastically reduce the
number of required values.

Secondly, efficient inference algorithms are formed in that work
by transmitting information between the local distributions rather
than working with the full joint distribution.

Thirdly, the separation of the qualitative representation of the
influences between variables from the numeric quantification of
the strength of the influences has a significant advantage for
knowledge engineering. When building a Bayesian network model, one
can focus first on specifying the qualitative structure of the
domain and then on quantifying the influences. When the model is
built, one is guaranteed to have a complete specification of the
joint probability distribution.

The most common computation performed on Bayesian networks is the
determination of the posterior probability of some random
variables, given the values of other variables in the network.
Because of the symmetric nature of conditional probability, this
computation can be used to perform both diagnosis and prediction.
Other common computations are:  the computation of the probability
of the conjunction of a set of random variables,  the computation
of the most likely combination of values of the random variables
in the network and  the computation of the piece of evidence that
has or will have the most influence on a given hypothesis.

A detailed discussion of inference techniques in Bayesian networks
can be found in the book by Pearl \cite{Pea00}.

\begin{itemize}

    \item \textbf{Probabilistic semantics.}
Any complete probabilistic model of a domain must, either
explicitly or implicitly, represent the joint distribution which
the probability of every possible event as defined by the values
of all the variables. There are exponentially many such events,
yet Bayesian networks achieve compactness by factoring the joint
distribution into local, conditional distributions for each
variable given its parents. If $x_{i}$ denotes some value of the
variable $X_{i}$ and $\pi(x_{i})$ denotes some set of values for
$X_{i}$'s parents $\pi(x_{i})$, then $P(x_{i}|\pi(x_{i}))$ denotes
this conditional distribution. For example, $P(x_{4}|x_{2},x_{3})$
is the probability of wetness given the values of sprinkler and
rain. Here $P(x_{4}|x_{2},x_{3})$ is the brief of
$P(x_{4}|\{x_{2},x_{3}\})$. The set parentheses are omitted for
the sake of readability. We use the same expression in this
thesis. The global semantics of Bayesian networks specifies that
the full joint distribution is given by the product

\begin{equation}\label{JPD}
P(x_{1},\ldots, x_{n})=\prod _{i}P(x_{i}|\pi(x_{i}))
\end{equation}

Equation \ref{JPD} is also called the chain rule for Bayesian networks.

\begin{figure}[htb]	
\centering
\includegraphics[width=5cm]{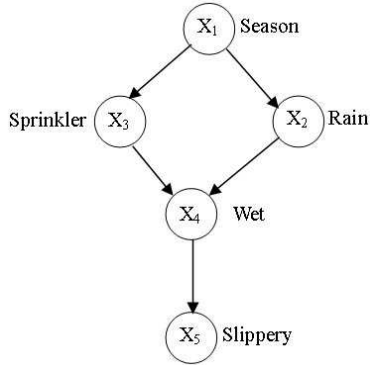} 
\caption{Causal Influences in A Bayesian Network.}\label{ExaBN}
\end{figure}

In the example Bayesian network in Figure \ref{ExaBN}, we have

\begin{equation}\label{JPD-Fig1}
P(x_{1}, x_{2}, x_{3}, x_{4},
x_{5})=P(x_{1})P(x_{2}|x_{1})P(x_{3}|x_{1})P(x_{4}|x_{2},x_{3})P(x_{5}|x_{4})
\end{equation}

Provided the number of parents of each node is bounded, it is easy
to see that the number of parameters required grows only linearly
with the size of the network, whereas the joint distribution
itself grows exponentially. Further savings can be achieved using
compact parametric representations, such as noisy-OR models,
decision tress, or neural networks, for the conditional
distributions \cite{Pea00}.

There are also entirely equivalent local semantics, which assert
that each variable is independent of its non-descendants in the
network given its parents. For example, the parents of $X_{4}$ in
Figure \ref{ExaBN} are $X_{2}$ and $X_{3}$ and they render $X_{4}$
independent of the remaining non-descendant, $X_{1}$. That is,

\begin{equation}
P(x_{4}|x_{1},x_{2},x_{3})=P(x_{4}|x_{2},x_{3})
\end{equation}

The collection of independence assertions formed in this way
suffices to derive the global assertion in Equation \ref{JPD-Fig1}, and vice
versa. The local semantics are most useful in constructing
Bayesian networks, because selecting as parents the direct causes
of a given variable automatically satisfies the local conditional
independence conditions. The global semantics lead directly to a
variety of algorithms for reasoning.

    \item \textbf{Evidential reasoning.}
From the product specification in Equation \ref{JPD-Fig1}, one can express
the probability of any desired proposition in terms of the
conditional probabilities specified in the network. For example,
the probability that the sprinkler was on, given that the pavement
is slippery, is

\begin{eqnarray}
&&P(X_{3}=on|X_{5}=true)\\ \nonumber
&&=\frac{P(X_{3}=on,X_{5}=true)}{P(X_{5}=true)}\\ \nonumber
&&=\frac{\sum_{x_{1},x_{2},x_{4}}P(x_{1},x_{2},X_{3}=on,x_{4},X_{5}=true)}{\sum_{x_{1},x_{2},x_{3},x_{4}}P(x_{1},x_{2},x_{3},x_{4},X_{5}=true)}\\
\nonumber
&&=\frac{\sum_{x_{1},x_{2},x_{4}}P(x_{1})P(x_{2}|x_{1})P(X_{3}=on|x_{1})P(x_{4}|x_{2},X_{3}=on)P(X_{5}=true|x_{4})}
 {\sum_{x_{1},x_{2},x_{3},x_{4}}P(x_{1})P(x_{2}|x_{1})P(x_{3}|x_{1})P(x_{4}|x_{2},x_{3})P(X_{5}=true|x_{4})}
\end{eqnarray}

These expressions can often be simplified in the
ways that reflect the structure of the network itself.

It is easy to show that reasoning in Bayesian networks subsumes
the satisfiability problem in propositional logic and hence
reasoning is NP-hard \cite{Coo90}. Monte Carlo simulation methods can
be used for approximate inference \cite{Pea87}, given that estimates
are gradually improved  as the sampling proceeds. (Unlike
join-tree methods, these methods use local message propagation on
the original network structure.) Alternatively, variational
methods \cite{JGJ98} provide bounds on the true probability.

    \item \textbf{Functional Bayesian networks.} The networks discussed so
far are capable of supporting reasoning about evidence and about
actions. Additional refinement is necessary in order to process
counterfactual information. For example, the probability that "the
pavement would not have been slippery had the sprinkler been
\textit{OFF}, given that the sprinkler is in fact \textit{ON} and that the
pavement is in fact slippery" cannot be computed from the
information provided in Figure \ref{ExaBN} and Equation \ref{JPD-Fig1}. Such
counterfactual probabilities require a specification in the form
of functional networks, where each conditional probability
$P(x_{i}|\pi(i))$ is replaced by a functional relationship
$x_{i}=f_{i}(\pi(i), \epsilon_{i})$, where $\epsilon_{i}$ is a
stochastic (unobserved) error term. When the functions $f_{i}$ and
the distributions of $\epsilon_{i}$ are known, all counterfactual
statements can be assigned unique probabilities, using evidence
propagation in a structure called a "twin network". When only
partial knowledge about the functional form of $f_{i}$ is
available, bounds can be computed on the probabilities of
counterfactual sentences \cite{BP95} \cite{Pea00}.

    \item \textbf{Causal discovery.} One of the most exciting prospects in
recent years has been the possibility of using Bayesian networks
to discover causal structures in raw statistical data \cite{PV91}
\cite{SGS93} \cite{Pea00}, which is a task previously considered impossible
without controlled experiments. Consider, for example, the
following pattern of dependencies among three events: $A$ and $B$
are dependent, $B$ and $C$ are dependent, yet $A$ and $C$ are
independent. If you ask a person to supply an example of three
such events, the example would invariably portray $A$ and $C$ as
two independent causes and $B$ as their common effect, namely,
$A\rightarrow B\leftarrow C$. Fitting this dependence pattern with
a scenario in which $B$ is the cause and $A$ and $C$ are the
effects is mathematically feasible but very unnatural, because it
must entail fine tuning of the probabilities involved; the desired
dependence pattern will be destroyed as soon as the probabilities
undergo a slight change.

Such thought experiments tell us that certain patterns of
dependency, which are totally void of temporal information, are
conceptually characteristic of certain causal directionalities and
not others. When put together systematically, such patterns can be
used to infer causal structures from raw data and to guarantee
that any alternative structure compatible with the data must be
less stable than the one(s) inferred; namely, slight fluctuations
in parameters will render that structure incompatible with the
data.

    \item \textbf{Plain beliefs.} In mundane decision making, beliefs are
revised not by adjusting numerical probabilities but by
tentatively accepting some sentences as "true for all practical
purposes". Such sentences, called plain beliefs, exhibit both
logical and probabilistic characters. As in classical logic, they
are propositional and deductively closed; as in probability, they
are subject to retraction and to varying degrees of entrenchment.
Bayesian networks can be adopted to model the dynamics of plain
beliefs by replacing ordinary probabilities with non-standard
probabilities, that is, probabilities that are infinitesimally
close to either zero or one \cite{GP96}.

    \item \textbf{Models of cognition.} Bayesian networks may be viewed as
normative cognitive models of propositional reasoning under
uncertainty \cite{Pea00}. They handle noise and partial information by
using local, distributed algorithm for inference and learning.
Unlike feed forward neural networks, they facilitate local
representations in which nodes correspond to propositions of
interest. Recent experiments \cite{TG01} suggest that they capture
accurately the causal inferences made by both children and adults.
Moreover, they capture patterns of reasoning that are not easily
handled by any competing computational model. They appear to have
many of the advantages of both the ``symbolic" and the
``subsymbolic" approaches to cognitive modelling.

Two major questions arise when we postulate Bayesian networks as
potential models of actual human cognition.

Firstly, does an architecture resembling that of Bayesian networks
exist anywhere in the human brain? No specific work had been done
to design neural plausible models that implement the required
functionality, although no obvious obstacles exist.

Secondly, how could Bayesian networks, which are purely
propositional in their expressive power, handle the kinds of
reasoning about individuals, relations, properties, and universals
that pervades human thought? One plausible answer is that Bayesian
networks containing propositions relevant to the current context
are constantly being assembled as needed to form a more
permanent store of knowledge. For example, the network in Figure \ref{ExaBN} may be assembled to help explain why this particular pavement
is slippery right now, and to decide whether this can be
prevented. The background store of knowledge includes general
models of pavements, sprinklers, slipping, rain, and so on; these
must be accessed and supplied with instance data to construct the
specific Bayesian network structure. The store of background
knowledge must utilize some representation that combines the
expressive power of first-order logical languages (such as
semantic networks) with the ability to handle uncertain
information.

   \end{itemize}

\section{Reasoning Structures in Bayesian Networks}

\subsection{Basic reasoning structures}

\subsubsection{d-Separation in Bayesian Networks}

d-Separation is one important property of Bayesian networks for
inference. Before we define d-separation, we first look at the way
that evidence is transmitted in Bayesian Networks. There are two
types of evidence:

\begin{itemize}
    \item \textbf{Hard Evidence} (instantiation) for a node $A$ is evidence
that the state of $A$ is definitely a particular value.
    \item \textbf{Soft Evidence} for a node $A$ is any evidence that enables
us to update the prior probability values for the states of $A$.
\end{itemize}

\textbf{d-Separation} (Definition):

Two distinct variables $X$ and $Z$ in a causal network are
d-separated if, for all paths between $X$ and $Z$, there is an
intermediate variable $V$ (distinct from $X$ and $Z$) such that
either

\begin{itemize}
    \item the connection is serial or diverging and $V$ is instantiated or
    \item the connection is converging, and neither $V$ nor any of $V$'s descendants have received evidence.
\end{itemize}

If $X$ and $Z$ are not d-separated, we call them d-connected.

\subsubsection{Basic structures of Bayesian Networks}

Based on the definition of d-seperation, three basic structures in Bayesian networks are as follows:

\begin{enumerate}

    \item \textbf{Serial connections}

Consider the situation in Figure \ref{serial-connection}. $X$ has an influence on $Y$,
which in turn has an influence on $Z$. Obviously, evidence on $Z$
will influence the certainty of $Y$, which then influences the
certainty of $Z$. Similarly, evidence on $Z$ will influence the
certainty on $X$ through $Y$. On the other hand, if the state of
$Y$ is known, then the channel is blocked, and $X$ and $Z$ become
independent. We say that $X$ and $Z$ are d-separated given $Y$,
and when the state of a variable is known, we say that it is
instantiated (hard evidence).

We conclude that evidence may be transmitted through a serial
connection unless the state of the variable in the connection is
known.

\begin{figure}[htb]	
\centering
\includegraphics[width=5cm]{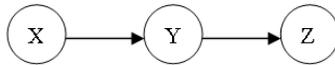} 
\caption{Serial Connection. When $Y$ is Instantiated, it blocks the communication between $X$ and $Z$.}\label{serial-connection}
\end{figure}

    \item \textbf{Diverging connections}

The situation in Figure \ref{Diverging-Connection} is called a diverging connection.
Influence can pass between all the children of $X$ unless the
state of $X$ is known. We say that $Y_{1}, Y_{2}, \ldots,Y_{n}$
are d-separated given $X$.

Evidence may be transmitted through a diverging connection unless
it is instantiated.

\begin{figure}[htb]	
\centering
\includegraphics[width=4cm]{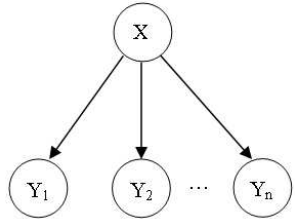} 
\caption{Diverging Connection. If $X$ is instantiated, it blocks the communication between its children.}\label{Diverging-Connection}
\end{figure}

    \item \textbf{Converging connections}

\begin{figure}[htb]	
\centering
\includegraphics[width=4cm]{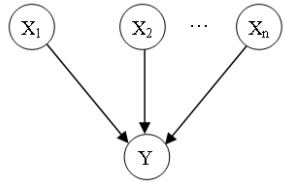} 
\caption{Converging Connection. If $Y$ changes certainty, it opens for the communication between its parents.}\label{Converging-Connection}
\end{figure}

A description of the situation in Figure \ref{Converging-Connection} requires a little
more care. If nothing is known about $Y$ except what may be
inferred from knowledge of its parents $X_{1}, \ldots,X_{n}$, then
the parents are independent: evidence on one of the possible
causes of an event does not tell us anything about other possible
causes. However, if anything is known about the consequences, then
information on one possible cause may tell us something about the
other causes.

This is the explaining away effect illustrated in
Figure \ref{ExaBN}. $X_{4}$ (pavement is wet) has occurred, and $X_{3}$
(the sprinkler is on) as well as $X_{2}$ (it's raining) may cause
$X_{4}$. If we then get the information that $X_{2}$ has occurred,
the certainty of $X_{3}$ will decrease. Likewise, if we get the
information that $X_{2}$ has not occurred, then the certainty of
$X_{3}$ will increase.

\end{enumerate}

The three preceding cases cover all ways in which evidence may be
transmitted through a variable.

\section{Classification of Inferences in Bayesian Networks}

In Bayesian networks, 4 popular inferences are identified as:

\begin{enumerate}
  \item Forward Inference

  Forward inferences is also called predictive inference (from causes to effects). The inference reasons from new information
about causes to new beliefs about effects, following the directions of the network
arcs. For example, in Figure \ref{serial-connection}, $X \rightarrow Y \rightarrow Z$ is a forward inference.

  \item Backward Inference

  Backward inferences is also called diagnostic inference (from effects to causes). The inference reasons from symptoms
to cause, Note that this reasoning occurs in
the opposite direction to the network arcs. In Figure \ref{serial-connection} , $Z \rightarrow Y$ is a backward inference. In Figure \ref{Diverging-Connection} , $Y_{i} \rightarrow X (i\in [1, n])$ is a backward inference.

  \item Intercausal Inference

  Intercausal inferences is also called \textbf{explaining away} (between parallel variables). The inference reasons about the mutual causes (effects) of a common
effect (cause). For example, in Figure \ref{Converging-Connection}, if the $Y$ is instantiated, $X_{i}$ and $X_{j} (i, j \in[1,n])$ are dependent. The reasoning $X_{i} \leftrightarrow X_{j} (i, j \in[1,n])$ is an intercausal inference. In Figure \ref{Diverging-Connection}, if $X$ is not instantiated, $Y_{i}$ and $Y_{j} (i, j \in[1,n])$ are dependent. The reasoning $Y_{i} \leftrightarrow Y_{j} (i, j \in[1,n])$  is an intercausal inference.

  \item Mixed inference

  Mixed inferences is also called combined inference. In complex Bayesian networks, the reasoning does not fit neatly into one of the types described above. Some inferences are a combination of several types of reasoning.

\end{enumerate}

\subsection{Inference in Bayesian Networks}

\subsubsection{inference in basic models}

\begin{itemize}
  \item in Serial Connections

    \begin{itemize}
      \item the \textbf{forward inference} executes with the evidence forward propagation. For example, in Figure \ref{SN-inference}, consider the inference $X \rightarrow Y \rightarrow Z$.
           \footnote{Note: In this chapter, $P(X^{+})$ is the abbreviation of $P(X=true)$, $P(X^{-})$ is the abbreviation of $P(|X=false)$. For simple expression, we use $P(Y|X)$ to denote $P(Y=true|X=true)$ by default. But in express  $P(Y^{+}|X)$, $X$ denotes both situations $X=true$ and $X=false$.}

\begin{figure}[htb]	
\centering
\includegraphics[width=8cm]{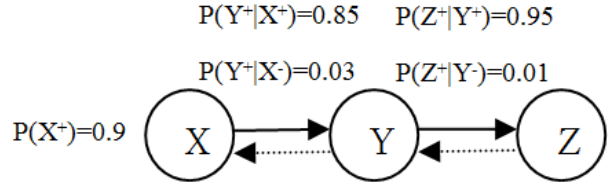} 
\caption{Inference in Serial Connection}
\label{SN-inference}
\end{figure}

          If Y is instantiated, X and Z are independent, then we have following example:

          $P(Z|XY)=P(Z|Y)$;

          $P(Z^{+}|Y^{+})=0.95$;

          $P(Z^{-}|Y^{+})=0.05$;

          $P(Z^{+}|Y^{-})=0.01$;

          $P(Z^{-}|Y^{-})=0.99$;

          if Y is not instantiated, X and Z are dependent, then

    $P(Z^{+}|X^{+}Y)=P(Z^{+}|Y^{+})P(Y^{+}|X^{+})+P(Z^{+}|Y^{-})P(Y^{-}|X^{+})$

    $=0.95*0.85+0.01*0.15=0.8075+0.0015=0.809$;

    $P(Z^{-}|X^{-}Y)=P(Z^{-}|Y^{+})P(Y^{+}|X^{-})+P(Z^{-}|Y^{-})P(Y^{-}|X^{-})$

    $=0.05*0.03+0.99*0.97=0.0015+0.9603=0.9618$.

      \item the \textbf{backward inference} executes the evidence backward propagation. For example, in Figure \ref{SN-inference}, consider the inference $Z \rightarrow Y \rightarrow X$.

          \begin{enumerate}
         \item  If $Y$ is instantiated ($P(Y^{+})=1$ or $P(Y^{-})=1)$, $X$ and $Z$ are independent, then

    \begin{eqnarray}\label{SN-backward-inference}
    P(X|YZ)=P(X|Y)=\frac{P(X)P(Y|X)}{P(Y)}
   \end{eqnarray}

   $P(X^{+}|Y^{+}Z)=P(X^{+}|Y^{+})=\frac{P(X^{+})P(Y^{+}|X^{+})}{P(Y^{+})}=\frac{09*0.85}{1}=0.765$;

   $P(X^{+}|Y^{-}Z)=P(X^{+}|Y^{-})=\frac{P(X^{+})P(Y^{-}|X^{+})}{P(Y^{-})}=\frac{09*0.15}{1}=0.135$.
   %

         \item
          If $Y$ is not instantiated, $X$ and $Z$ are dependent (See the dashed lines in Figure \ref{SN-inference}).
          Suppose $P(Z^{+})=1$ then

$P(X^{+}|YZ^{+})=\frac{P(X^{+}YZ^{+})}{P(YZ^{+})}=\frac{P(X^{+}YZ^{+})}{\sum_{X}P(XYZ^{+})}$;\\

$P(X^{+}YZ^{+})=P(X^{+}Y^{+}Z^{+})+P(X^{+}Y^{-}Z^{+})=0.9*0.85*0.95+0.9*0.15*0.05=0.72675+0.00675=0.7335$;\\

$\sum_{X}P(XYZ^{+})=P(X^{+}Y^{+}Z^{+})+P(X^{+}Y^{-}Z^{+})+P(X^{-}Y^{+}Z^{+})+P(X^{-}Y^{-}Z^{+})\\
=0.9*0.85*0.95+0.9*0.15*0.99+0.1*0.03*0.95+0.1*0.97*0.01\\
=0.72675+0.13365+0.00285+0.00097=0.86422$;\\

$P(X^{+}|YZ^{+})=\frac{P(X^{+}YZ^{+})}{\sum_{X}P(XYZ^{+})}=\frac{0.7335}{0.86422}=0.8487.$

\end{enumerate}
  \end{itemize}
     In serial connections, there is no intercausal inference.

  \item in Diverging Connections

    \begin{itemize}
      \item the \textbf{forward inference} executes with the evidence forward propagation. For example, in Figure \ref{DN-inference}, consider the inference $Y \rightarrow X$ and $Y \rightarrow Z$, the goals are easy to obtain by nature.

          \begin{figure}[htb]	
\centering
\includegraphics[width=7cm]{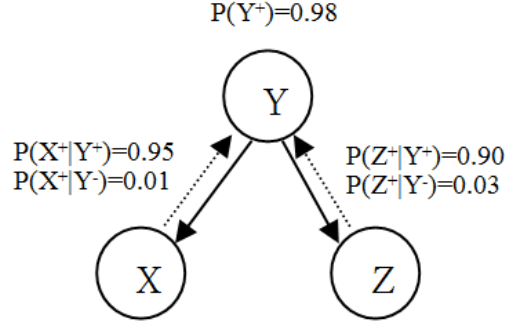} 
\caption{Inference in Diverging Connection}
\label{DN-inference}
\end{figure}

      \item the \textbf{backward inference} executes with the evidence backward propagation, see the dashed line in Figure \ref{DN-inference}, consider the inference $(XZ) \rightarrow Y$, $X$ and $Z$ are instantiated by assumption, suppose $P(X^{+}=1)$, $P(Z^{+}=1)$. Then,
   \begin{eqnarray} \label{DN-backward-inference}
          P(Y^{+}|X^{+}Z^{+})=\frac{P(Y^{+}X^{+}Z^{+})}{P(X^{+}Z^{+})}=\frac{P(Y^{+})P(X^{+}|Y^{+})P(Z^{+}|Y^{+})}{P(X^{+}Z^{+})} \nonumber\\
          =\frac{0.98*0.95*0.90}{1}=0.8379
          \end{eqnarray}
      \item the intercausal inference executes between effects with a common cause. In Figure \ref{DN-inference},
      if $Y$ is not instantiated, there exists intercausal inference in diverging connections.
      Consider the inference $X \rightarrow Z$,

      $P(X^{+}|YZ^{+})=\frac{P(X^{+}YZ^{+})}{P(YZ^{+})}=\frac{P(X^{+}Y^{+}Z^{+})+P(X^{+}Y^{-}Z^{+})}{P(Y^{+}Z^{+})+P(Y^{-}Z^{+})}$;

      $=\frac{0.98*0.95*0.90+0.02*0.01*0.03}{0.98*0.90+0.02*0.03}=0.94936$.
%
    \end{itemize}

  \item in Converging Connections,

    \begin{itemize}
      \item the \textbf{forward inference} executes with the evidence forward propagation. For example, in Figure \ref{CN-inference}, consider the inference $(XZ) \rightarrow Y$, $P(Y|XZ)$ is easy to obtain by the definition of Bayesian Network in by nature.

           \begin{figure}[htb]	
\centering
\includegraphics[width=7cm]{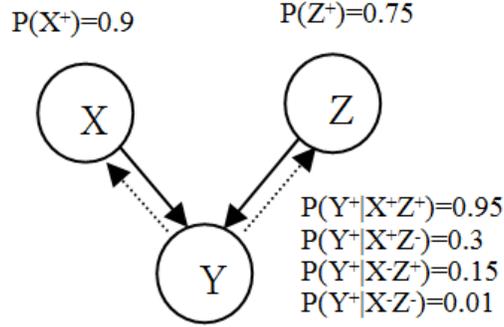} 
\caption{Inference in Converging Connection}
\label{CN-inference}
\end{figure}

      \item the \textbf{backward inference} executes with the evidence backward propagation. For example, in Figure \ref{CN-inference}, consider the inference $Y \rightarrow (XZ)$.

          $P(Y)=\sum_{XZ}P(XYZ)=\sum_{XZ}(P(Y|XZ)P(XZ))$,\\

          $P(XZ|Y)=\frac{P(Y|XZ)P(XZ)}{P(Y)}=\frac{P(Y|XZ)P(X)P(Z)}{\sum_{XZ}(P(Y|XZ)P(XZ))}$.\\

          Finally,

          $P(X|Y)=\sum_{Z}P(XZ|Y)$,

          $P(Z|Y)=\sum_{X}P(XZ|Y)$.

      \item the \textbf{intercausal inference} executes between causes with a common effect, and the intermediate node is instantiated, then $P(Y^{+})=1$ or $P(Y^{-})=1$. In Figure \ref{CN-inference}, consider the inference $X \rightarrow Z$, suppose $P(Y^{+})=1$,

          $P(Z^{+}|X^{+}Y^{+})=\frac{P(Z^{+}X^{+}Y^{+})}{P(X^{+}Y^{+})}=\frac{P(Z^{+}X^{+}Y^{+})}{\sum_{Z}P(X^{+}Y^{+}Z)}$;\\

         $P(Z^{+}X^{+}Y^{+})=P(X^{+})P(Z^{+})P(Y^{+}|X^{+}Z^{+})$;\\

          $\sum_{Z}P(X^{+}YZ)=P(X^{+}Y^{+}Z^{+})+P(X^{+}Y^{+}Z^{-})$;\\

          $P(Z^{+}|X^{+}Y^{+})=\frac{P(Z^{+}X^{+}Y^{+})}{\sum_{Z}P(X^{+}Y^{+}Z)}=\frac{P(X^{+})P(Z^{+})P(Y^{+}|X^{+}Z^{+})}{P(X^{+}Y^{+}Z^{+})+P(X^{+}Y^{+}Z^{-})}.$

    \end{itemize}
\end{itemize}

\subsubsection{inference in complex model}

For complex models in Bayesian networks, there are single-connected networks, multiple-connected, or event looped networks. It is possible to use some methods, such as Triangulated Graphs, Clustering and Join Trees \cite{BB72} \cite{FT07} \cite{Gol80}, etc., to simplify them into a polytree. Once a polytree is obtained, the inference can be executed by the following approaches.

Polytrees have at most one path between any pair of nodes; hence they are also referred to as singly-connected networks.

\begin{figure}[htb]	
\centering
\includegraphics[width=5cm]{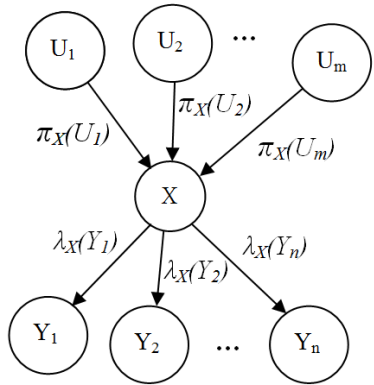} 
\caption{Evidence Propagation in Polytree}
\label{polytree}
\end{figure}

Suppose $X$ is the query node, and there is some set of evident nodes $E, X \notin E$. The posterior probability (belief) is denoted as $\mathbb{B}(X)=P(X|E)$, see Figure \ref{polytree}.

$E$ can be splitted into 2 parts: $E^{+}$ and $E^{-}$. $E^{-}$ is the part consisting of assignments to variables in the subtree rooted at $X$, $E^{+}$ is the rest of it.

$\pi_{X}(E^{+})=P(X|E^{+})$

$\lambda_{X}(E^{-})=P(E^{-}|X)$
\begin{equation}\label{Belief}
    \mathbb{B}(X)=P(X|E)=P(X|E^{+}E^{-})=\frac{P(E^{-}|XE^{+})P(X|E^{+})}{P(E^{-}|E^{+})}
=\frac{P(E^{-}|X)P(X|E^{+})}{P(E^{-}|E^{+})}=\alpha\pi_{X}(E^{+})\lambda_{X}(E^{-})
\end{equation}

$\alpha$ is a constant independent of $X$.

where

\begin{equation}\label{lamda}
    \lambda_{X}(E^{-})=\{\begin{array}{cc}
                     1 &  if\ evidence\ is\ X=x_{i} \\
                     0 & if\ evidence\ is\ for\ another\ x_{j} \\
                   \end{array}
\end{equation}

\begin{equation}\label{Pi}
    \pi_{X}(E^{+})=\sum_{u_{1},...,u_{m}}P(X|u_{1},...,u_{m})\prod_{i}\pi_{X}(u_{i})
\end{equation}

\begin{enumerate}
  \item Forward inference in Polytree

  Node $X$ sends $\pi$ messages to its children.
  \begin{equation}\label{Forward-inference-polytree}
    \pi_{X}(U)=\{\begin{array}{cc}
                        1 & if\ x_{i}\in X\ is\ entered \\
                        0 & if\ evidentce\ is\ for\ another\ value\ x_{j} \\
                        \sum_{u_{1},...u_{m}}P(X|u_{1},...u_{m})\prod_{i}\pi_{X}(u_{i}) & otherwise
                      \end{array}
  \end{equation}

  \item Backward inference in Polytree
Node $X$ sends new $\lambda$ messages to its parents.
\begin{equation}\label{backward-inference-polytree}
    \lambda_{X}(Y)=\prod_{y_{j}\in Y}[\sum_{j}P(y_{j}|X)\lambda_{X}(y_{j})]
\end{equation}

\end{enumerate}

\subsection{Related Algorithms for Probabilistic Inference}

Various types of inference algorithms exist for Bayesian networks \cite{LS88} \cite{Pea88} \cite{Pea00} \cite{Nea93}.
Each class offers different properties and works better
on different classes of problems, but it is very unlikely that a
single algorithm can solve all possible problem instances
effectively. Every resolution is always based on a particular
requirement. It is true that almost all computational problems and
probabilistic inference using general Bayesian networks have been
shown to be NP-hard by Cooper \cite{Coo90}.

In the early 1980's, Pearl published an efficient message
propagation inference algorithm for polytrees \cite{KP83} \cite{Pea86}. The
algorithm is exact, and has polynomial complexity in the number of
nodes, but works only for singly connected networks. Pearl also
presented an exact inference algorithm for multiple connected
networks called loop cutset conditioning algorithm \cite{Pea86}. The
loop cutset conditioning algorithm changes the connectivity of a
network and renders it singly connected by instantiating a
selected subset of nodes referred to as a loop cutset. The
resulting single connected network is solved by the polytree
algorithm, and then the results of each instantiation are weighted
by their prior probabilities. The complexity of this algorithm
results from the number of different instantiations that must be
considered. This implies that the complexity grows exponentially
with the size of the loop cutest being $O(d^{c})$, where $d$ is
the number of values that the random variables can take, and $c$
is the size of the loop cutset. It is thus important to minimize
the size of the loop cutset for a multiple connected network.
Unfortunately, the loop cutset minimization problem is NP-hard. A
straightforward application of Pearl's algorithm to an acyclic
digraph comprising one or more loops invariably leads to
insuperable problems \cite{KW01} \cite{Nea93}.

Another popular exact Bayesian network inference algorithm is
Lauritzen and Spiegelhalter's clique-tree propagation algorithm
\cite{LS88}. It is also called a "clustering" algorithm. It first
transforms a multiple connected network into a clique tree by
clustering the triangulated moral graph of the underlying
undirected graph and then performs message propagation over the
clique tree. The clique propagation algorithm works efficiently
for sparse networks, but still can be extremely slow for dense
networks. Its complexity is exponential in the size of the largest
clique of the transformed undirected graph.

In general, the existent exact Bayesian network inference
algorithms share the property of run time exponentiality in the size
of the largest clique of the triangulated moral graph, which is
also called the induced width of the graph \cite{LS88}. 
%

\section{Conclusion}

This chapter summarizes the popular inferences methods in Bayesian networks. The results demonstrates that the evidence can propagated across the Bayesian networks by any links, whatever it is forward or backward or intercausal style. The belief updating of Bayesian networks can be obtained by various available inference techniques. Theoretically, exact inferences in Bayesian networks is feasible and manageable. However, the computing and inference is NP-hard. That means, in applications, in complex huge Bayesian networks, the computing and inferences should be dealt with strategically and make them tractable. Simplifying the Bayesian networks in structures, pruning unrelated nodes, merging computing, and approximate approaches might be helpful in the inferences of large scale Bayeisan networks.

\end{document}